\documentclass{article}

\usepackage[preprint]{neurips_2023}

\usepackage[utf8]{inputenc} % allow utf-8 input
\usepackage[T1]{fontenc}    % use 8-bit T1 fonts
\usepackage{hyperref}       % hyperlinks
\usepackage{url}            % simple URL typesetting
\usepackage{booktabs}       % professional-quality tables
\usepackage{amsfonts}       % blackboard math symbols
\usepackage{nicefrac}       % compact symbols for 1/2, etc.
\usepackage{microtype}      % microtypography
\usepackage{xcolor}         % colors
\usepackage{graphicx}
\usepackage{amsmath}
\usepackage{multirow}

\usepackage{caption}
\title{Motion Manipulation via Unsupervised Keypoint Positioning in Face Animation}

\author{%
  Hong Li$^{1,}$\thanks{Equal contribution.} \\
  BUAA \\
  \texttt{link0502@buaa.edu.cn} \\
  \And
  Boyu Liu$^{1,}$\footnotemark[1] \\
  BUAA \\
  \texttt{byliu@buaa.edu.cn} \\
  \And
  Xuhui Liu$^{2}$ \\
  KAUST \\
  \texttt{xhliu.comp@gmail.com} \\
  \And
  Baochang Zhang$^{1}$\thanks{Corresponding author.} \\
  BUAA \\
  \texttt{bczhang@buaa.edu.cn}
}

\begin{document}

\maketitle

%-------------------------------------------------------------------------
\begin{abstract}
Face animation deals with controlling and generating facial features with a wide range of applications. The methods based on unsupervised keypoint positioning can produce realistic and detailed virtual portraits. However, they cannot achieve controllable face generation since the existing keypoint decomposition pipelines fail to fully decouple identity semantics and intertwined motion information (e.g., rotation, translation, and expression). To address these issues, we present a new method, Motion Manipulation via unsupervised keypoint positioning in Face Animation (MMFA). We first introduce self-supervised representation learning to encode and decode expressions in the latent feature space and decouple them from other motion information. Secondly, we propose a new way to compute keypoints aiming to achieve arbitrary motion control. Moreover, we design a variational autoencoder to map expression features to a continuous Gaussian distribution, allowing us for the first time to interpolate facial expressions in an unsupervised framework. We have conducted extensive experiments on publicly available datasets to validate the effectiveness of MMFA, which show that MMFA offers pronounced advantages over prior arts in creating realistic animation and manipulating face motion.
\end{abstract}

\section{Introduction}
Face animation aims to generate a photo-realistic continuous face motion video from static images with movement information provided by a sequence of video frames.
The synthesis of high-fidelity facial animation can enhance user experience in various applications, including remote consoles, video conferencing, and online customer service. Nevertheless, ensuring movement consistency while minimizing variance in facial identity poses a significant and complex challenge.

The mainstream works focus on constraining facial geometry and kinematics through the use of 2D landmark correspondences \cite{yang2022face2face, geng2018warp}, 3D morphable model (3DMM) coefficients \cite{ren2021pirenderer, doukas2021headgan, yin2022styleheat} and other priors. Nevertheless, due to the insufficiency of modeling facial details using existing priors, face animation remains challenging in accurately capturing expression changes. Several methods \cite{wiles2018x2face, siarohin2019first, wang2021one, siarohin2021motion, zeng2022fnevr, zeng2023face} adopt a warping strategy for image deformation by estimating the motion optical flow via unsupervised keypoint positioning, which minimizes the reliance on prior information and enhances the model's generalization capacity. But they fail to decouple identity semantics from motion information, forsaking the separate manipulation of facial movement. 
Recent methods \cite{wang2021one} have explored face modeling with unsupervised keypoint positioning, assuming faces can be represented by 3D keypoints to enable arbitrary alteration over pose attributes, such as rotation and translation. However, the expression represented by these keypoints' deformation exhibits significant coupling with facial scaling, limiting accurate expression manipulation.
Rather than directly modifying images, some methods \cite{wang2022latent, oorloff2023one, ArunMallya2022ImplicitWF} transfer motion from driving frames to static images by replacing or linearly transforming latent vectors. Among them, DPE \cite{pang2023dpe} tries to decouple the facial features using an unsupervised method. Yet they fail to explicitly model facial expressions or poses.

\begin{figure}[t]
\begin{center}
   \includegraphics[width=\linewidth]{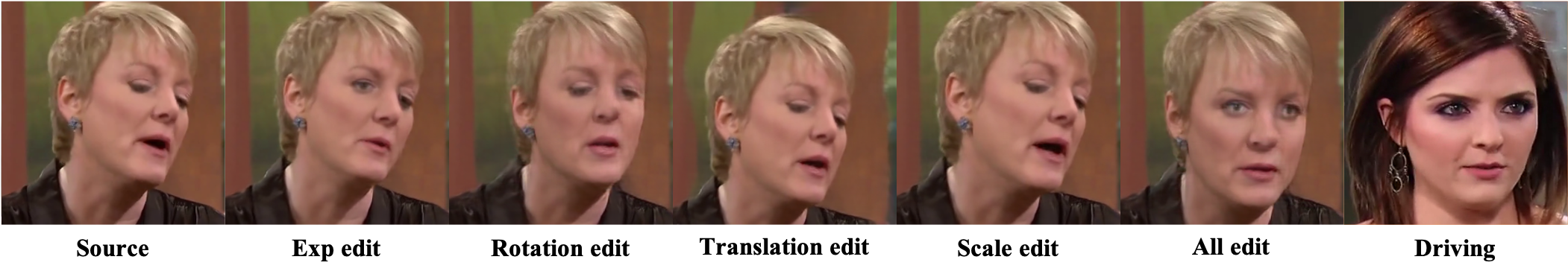}
\end{center}
   \caption{An example of MMFA. MMFA can realize realistic motion attribute editing while achieving face animation.}
\label{fig:introduction}
\end{figure}

In this paper, we enable face animation with self-supervised representation learning to effectively control the facial attributes, while leveraging unsupervised keypoint positioning to maintain the transfer ability of the motion details. We present a new face animation network that explicitly achieves motion manipulation, named MMFA. An example is shown in Fig. \ref{fig:introduction}.
Specifically, we design a new keypoint decomposition pipeline based on a scaled orthographic projection, where a scaling factor is estimated to handle the perspective effect caused by the distance variation between the face and the camera.
Grounded on this pipeline, to achieve robust face motion control, an encoder-decoder model with self-supervised representation learning is employed to extract corresponding expression latent features, which are then decoupled from other attributes, such as scaling, translation, and rotation.
To fully explore the latent space of expression, we also employ a variational autoencoder (VAE) \cite{kingma2013auto} to map facial expression information from images to a continuous latent space. This simple VAE with the adversarial loss enables reconstructing expression latent features for continuous control.
By learning this mapping, our MMFA generates user-desired virtual animation while maintaining consistency in the video, enabling natural human-computer interaction.

In summary, the main contributions of this paper are as follows:
\begin{itemize}
\item 
We propose MMFA based on unsupervised keypoint positioning and self-supervised representation learning. The pose and expression are efficiently decoupled with reasonable assumptions, achieving precise attribute control with minimal priors.

\item 
We leverage a VAE to map the expression feature to a continuous latent space for expression manipulation.
The Gaussian distribution provides a new means of relating facial expression interpolation for coherent animation and expands the application scenarios of the model.

\item We conduct extensive experiments to validate the effectiveness of our method over prior arts. The lowest FID shows that MMFA has a great advantage in the authenticity of generation, and has the ability to efficiently transfer facial details.
\end{itemize}

%------------------------------------------------------------------------
\section{Related Work}
%-------------------------------------------------------------------------
\subsection{Face Animation}
\textbf{Keypoint-Based Methods} 
% Leverage sparse keypoints to determine facial motion fields for manipulating target images without relying on facial priors. Monkey-Net first uses abstract keypoints centered on faces/bodies for image deformation. FOMM improves Monkey-Net's motion estimation with Jacobians, enabling better motion consistency.
leverage sparse keypoints to establish the motion field for manipulating face movement in the target image without relying on facial priors \cite{siarohin2019animating, siarohin2019first, wang2021one, siarohin2021motion, wang2021safa, zeng2022fnevr, ArunMallya2022ImplicitWF}. Monkey-Net \cite{siarohin2019animating} first uses abstract keypoints centered on faces/bodies for image deformation. FOMM \cite{siarohin2019first} improves Monkey-Net's motion estimation with Jacobians, enabling better motion consistency. Face-vid2vid \cite{wang2021one} proposes a 3D facial keypoints estimation pipeline, which introduces control parameters such as rotation and translation to realize simple face pose editing.
% Due to FOMM's strong motion detail transfer, MRAA \cite{siarohin2021motion}  designs a PCA-based motion estimator for articulated object animation. DaGAN \cite{hong2022depth} use 3D geometric information to guide keypoints prediction by means of depth estimation. 
Despite the fact that the estimated keypoints do not embody explicit semantics, extensive experiments \cite{jakab2018unsupervised, siarohin2019animating, siarohin2019first} show that this strategy is capable of learning effective geometric representations for detailed motion transfer. 
However, lacking priors, these methods cannot arbitrarily control facial features and may cause identity leakage in cross-identity reenactment.

\textbf{3DMM-Based Methods.} By leveraging facial priors of 3DMMs \cite{paysan20093d, gerig2018morphable,bolkart2015groupwise, bolkart2016robust,booth20163d, li2017learning}, several methods \cite{thies2016face2face, doukas2021headgan, ren2021pirenderer, wang2021safa, yin2022styleheat, yang2022face2face} use 3D geometric information to guide motion field estimation and make it possible to decouple and control facial features in face animation while preserving 3D consistency. HeadGAN \cite{doukas2021headgan} and SAFA \cite{wang2021safa} use the 3D face rendering results based on 3DMM fitting to estimate the facial motion optical flow, which can well maintain the identity of the generated results. Instead of using explicit facial priors, PIRenderer \cite{ren2021pirenderer} and StyleHEAT \cite{yin2022styleheat} explore the calculation method of flow fields using 3DMM coefficients directly as motion representation, which simplifies the estimation network. % Furthermore, Face2face Rho \cite{yang2022face2face} uses 3DMM to estimate the 2D sparse landmarks of the face for motion estimation, which greatly improves the inference speed.
The above methods have a significant advantage in the authenticity of face generation due to facial priors, but since most 3DMMs are based on linear bases, the ability to transfer details by these models is limited.

\textbf{Latent Space-Based Methods.} In order to better encode facial features and achieve separate motion control, several methods \cite{wang2022latent, pang2023dpe} directly learn the latent space of facial features and accomplish the decoupling of expression and pose. Besides, recent researches \cite{tewari2020stylerig, ghosh2020gif, gu2021stylenerf} show that StyleGANs \cite{karras2019style, karras2020analyzing} can serve as a promising prior for downstream tasks on image synthesis.
% With StyleGAN's latent feature space, StyleHEAT \cite{yin2022styleheat} provides high-resolution image generation and attribute editing in face animation via GAN inversion.
A hybrid latent-space is employed in \cite{oorloff2023one} with StyleGAN2 \cite{karras2020analyzing}, encoding a given image into identity latent and facial deformation latent for high-fidelity face animation and semantic editing. 
All StyleGAN-based methods of face animation are undoubtedly constrained by prior information from the pre-trained models and show deficiencies in the detailed manipulation of expressions.

\begin{figure}[t]
\begin{center}
   \includegraphics[width=\linewidth]{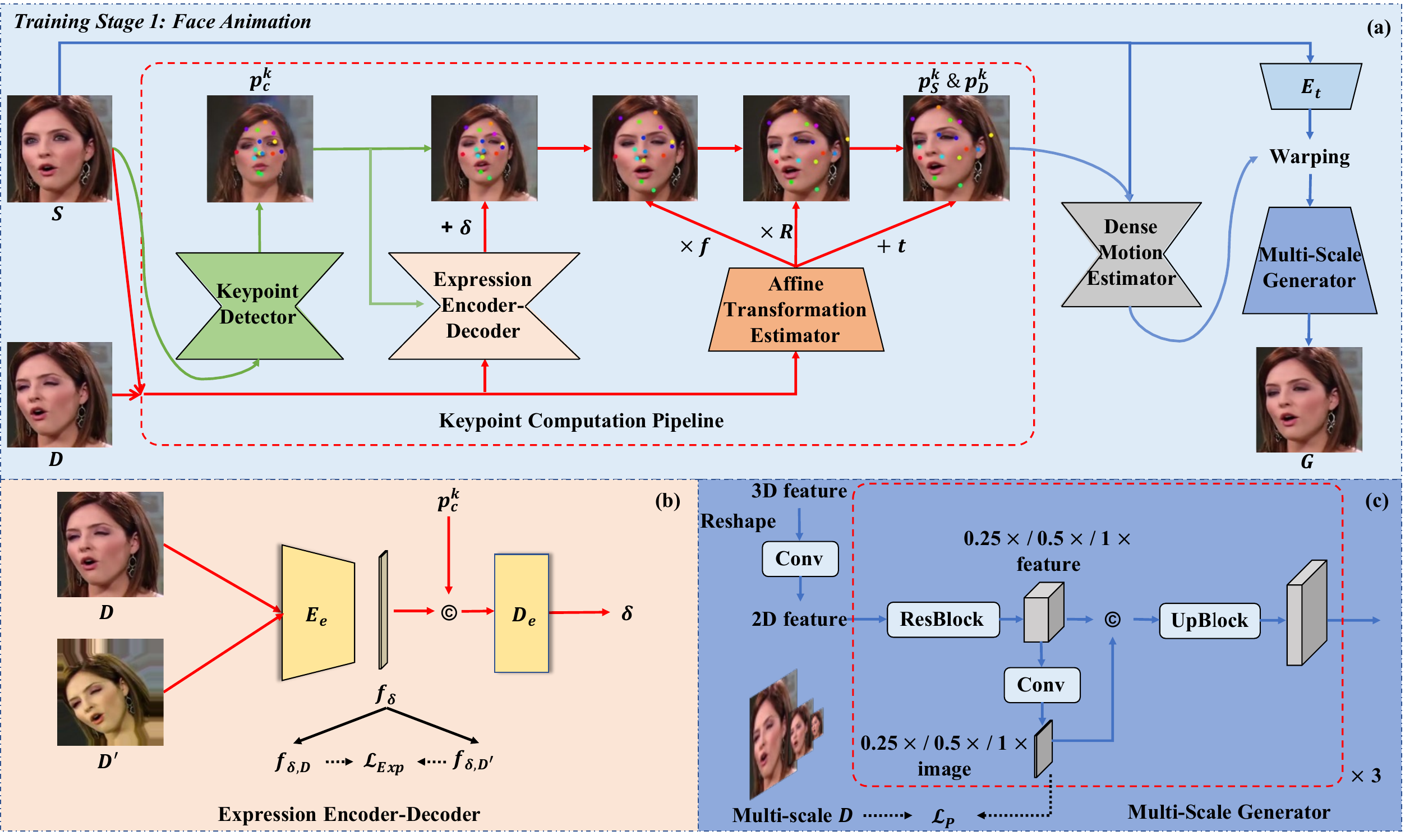}
\end{center}
   \caption{Overview of MMFA, where $\copyright$ indicates channel-wise concatenation. Part (a) details the steps by which our model decomposes keypoints. Part (b) describes the expression encoder-decoder structure and self-supervised representation learning. Part (c) shows the structure of the multi-scale generator.}
\label{fig:main}
\end{figure}

%-------------------------------------------------------------------------
% \subsection{Contrastive learning}
% Contrastive learning \cite{hadsell2006dimensionality} is a commonly used data encoding method in self-supervised and unsupervised learning, which often uses contrastive losses to measure the similarities between a pair of data in a representation space. Memory bank \cite{wu2018unsupervised} is an effective way to use negative samples to achieve contrastive learning, but its training can be intractable on large-scale datasets. Furthermore, MoCo \cite{he2020momentum} proposed a more memory-efficient strategy, maintaining a queue of negative samples with momentum-updated encoder. SimCLR \cite{chen2020simple} is a simple contrastive learning method, which only performs data augmentations on samples in the same batch, without requiring specialized architectures or a memory bank.
% Moreover, SimSiam \cite{chen2021exploring} achieves effective contrastive learning without using negative samples.
\section{Method}
This work is a 3D keypoint-based face animation approach. To enable independent editing of pose and expression, we adopt a keypoint decomposition framework based on self-supervised representation learning. We also train a latent space in a VAE to model facial expressions independently, thereby achieving expression control.
Fig. \ref{fig:main} illustrates the complete pipeline.
% We begin by revising the previous keypoint-based framework in Sec. \ref{rethinking}. We then present our improved framework in Sec. \ref{pipeline}, followed by details on the latent space VAE for facial expressions in Sec. \ref{vae}.

%-------------------------------------------------------------------------
\subsection{Revising the Keypoint-Based Framework} \label{rethinking}
We formulate face animation as face synthesis frame by frame, in which a source face image $S$ and a driving face video sequence with $N$ frames $\{D_{1}, D_{2}, ..., D_{N}\}$ are given, where $D_{i}$ represents the $i$-th frame in the video. The objective of generation is to extract the facial motion from the driving frame $D_{i}$ and generate a virtual face, which has the same pose and expression as the driving frame $D_{i}$ while maintaining the identity of $S$.

As the most representative method based on unsupervised keypoint positioning, FOMM \cite{siarohin2019first} estimates a set of $K$ keypoints $\{{p_{S}^{k}, p_{D}^{k} \in \mathbb{R}^2}\}$ and corresponding Jacobians $\{J_{S}^{k}, J_{D}^{k} \in \mathbb{R}^{2 \times 2}\}$ from $S$ and $D$ respectively to depict the transformation of pixel $z \in \mathbb{R}$ in the vicinity of each keypoint by an affine motion function:
\begin{equation}
\small
  \begin{aligned}
  A_{S \gets D}^{k}(z) &= p_{S}^{k} + J^{k}(z - p_{D}^{k}), \\
  \end{aligned}
  \label{sparse motion}
\end{equation}
where $J^{k}=J_{S}^{k}(J_{D}^{k})^{-1}$. By weighting the above $K$ sparse motions and static background, the desired dense backward optical flow from $D$ to $S$ can be calculated as:
\begin{equation}
\small
  \begin{aligned}
  \mathcal{T}_{S \gets D}(z) &= M^{0} z + \sum^{K}_{k=1} M^{k} A_{S \gets D}^{k}(z), \\
  \end{aligned}
\end{equation}
where $M^{k}$ are motion weights predicted by a U-Net \cite{ronneberger2015u}.

% After applying 2D warping to the source feature $F_{S}$ encoded from $S$, the deformed feature $F_{S, warped}$ is finally processed by a subsequent decoder to render virtual face image $G$. FOMM provides an effective framework for face animation. However, since there are no constraints on the keypoints, it is easy to cause the identity changing of the source face during the cross-identity expression transfer.

In order to achieve simple head pose control, Face-vid2vid \cite{wang2021one} extends the keypoints of FOMM to 3D space and proposes an intuitive decomposition pipeline. Instead of finding the keypoints that correspond to the face, the new keypoint estimator predicts a set of $K$ points $\{{p_{C}^{k} \in \mathbb{R}^2}\}$ representing the source face in a neutral pose and expression, namely canonical keypoints. Subsequently, the rotation matrices $\{R_{S}, R_{D} \in \mathbb{R}^{3 \times 3}\}$, translation vectors $\{t_{S}, t_{D} \in \mathbb{R}^{3}\}$ and expression deformations $\{\delta^{k}_{S}, \delta^{k}_{D} \in \mathbb{R}^{3}\}$ of $S$ and $D$ are estimated separately through a pose-expression network. The 3D keypoints $\{{p_{S}^{k}, p_{D}^{k} \in \mathbb{R}^2}\}$ required for dense optical flow estimation can be obtained by transforming the canonical keypoints as:
\begin{equation}
\small
  \begin{aligned}
  p_{S}^{k} = R_{S}p_{C}^{k} + t_{S} + \delta_{S}^{k}, ~p_{D}^{k} = R_{D}p_{C}^{k} + t_{D} + \delta_{D}^{k}. \\
  \end{aligned}
  \label{source kp}
\end{equation}

% Finally, we perform the warping of the source feature $F_{S}$ in 3D and use the rotation matrix $R_{S}, R_{D}$ to replace the Jacobian to generate the virtual image.

\begin{figure}[t]
\begin{center}
   \includegraphics[width=\linewidth]{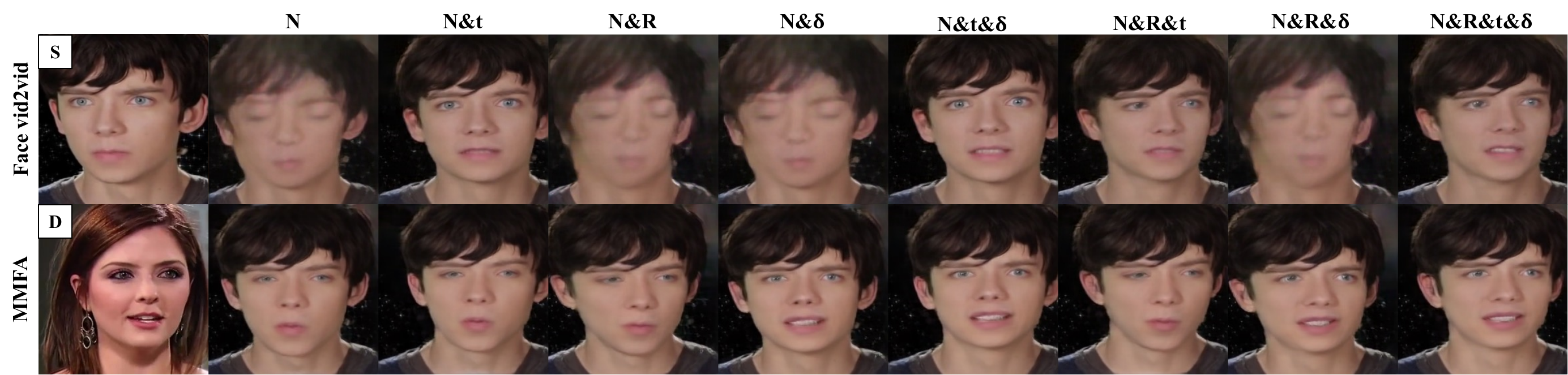}
\end{center}
   % \caption{Visual comparison of attribute editing between MMFA and Face vid2vid. The second column displays the semantic images corresponding to the neutral keypoints defined as keypoints with no rotation, translation and expression correction. It is important to note that in MMFA, the neutral keypoints have been scaled according to the canonical keypoints, while in Face vid2vid, the canonical keypoints are also the neutral keypoints. Subsequent columns show the semantic images obtained after applying translations, rotations, expression deformations, translation and expression deformations, rotation and expression deformations, and rotation, translation, and expression deformations, respectively.}
    \caption{Visual comparison of attribute editing between MMFA and Face vid2vid. The second column displays the semantic images corresponding to the neutral keypoints $N$ defined as keypoints with no rotation, translation, and expression correction. It is important to note that in MMFA, the neutral keypoints have been scaled according to the canonical keypoints, while in Face vid2vid, the canonical keypoints are also the neutral keypoints. Subsequent columns show the semantic images obtained after applying translations $t$, rotations $R$, and expression deformations $\delta$.}
\label{fig:motivation}
\end{figure}

While acknowledging the notable improvement of generation quality by Face-vid2vid, we find that, in the commonly used training data \cite{nagrani2017voxceleb, chung2018voxceleb2}, faces from different frames in the same video are located in diverse 3D positions, 
% When the speaker moves along the optical axis in the camera coordinate system, there may be a non-negligible difference in the scale of the faces from two frames selected as source image $S$ and driving image $D$.
which causes unavoidable face scale inconsistency. However, since Face-vid2vid ignores the perspective projection of the camera, 
% only rotation $R$ and translation $t$ cannot adjust the scale of the keypoints' closure
it forces the expression deformation $\delta$ to learn the change of face shape. % Moreover, as shown in \ref{fig:motivation}, we also observed an obvious coupling between expression deformation and rotation through experiments, which hinders our use of unsupervised keypoint positioning to manipulate expressions.
% Moreover, as shown in \ref{fig:motivation}, previous keypoint-based methods has incapability of fully decoupling identity semantics and motion information, which hinders our use of unsupervised keypoint positioning to manipulate face movement.
Moreover, as shown in Fig. \ref{fig:motivation}, Face-vid2vid has the incapability of fully decoupling identity semantics and motion information, which hinders the use of unsupervised keypoint positioning to manipulate face movement.

%-------------------------------------------------------------------------
\subsection{Pipline of MMFA} \label{pipeline}
We first introduce three assumptions regarding the face to facilitate the subsequent keypoint transformations: 

\textbf{Assumption 1.} The centroid of the object lies at the origin of the world coordinate frame, which coincides with the camera coordinate frame. 

\textbf{Assumption 2.} The mapping from the camera coordinate frame to the image coordinate frame adheres to the orthographic projection. 

\textbf{Assumption 3.} The object is assumed to be rigid. When scale and rotational transforms are applied to the face, the topological relationships between regions remain invariant.

Assumption 1 and Assumption 3 allow us to use the rotation matrix and translation directly to change the facial pose. Assumption 2 converts the perspective projection into the orthographic projection while making the scale of the face independent of the position of the optical axis, which simplifies the control of the face in 3D. Referring to the face description of 3DDFA \cite{zhu2017face}, we decompose the modeling of the face keypoints into $K$ canonical keypoints $\{{p_{C}^{k} \in \mathbb{R}^3}\}$, rotation $R \in \mathbb{R}^{3 \times 3}$, translation $t \in \mathbb{R}^{2}$, scale $f \in \mathbb{R}$, and expression deformations $\{\delta^{k} \in \mathbb{R}^{3}\}$. Among them, the canonical keypoints represent the spatial anchors of the identity.

We use the same keypoint detector in \cite{wang2021one} with a 2D encoder and a 3D decoder to predict $p_{C}^{k}$. Besides, we employ similar affine transformation estimators to obtain the parameters $R, f$, and $t$. Specifically, a simple decoder is used to learn the scale $f$ and the transformation $t$. To avoid the additional influence of the expression deformation $\delta^{k}$ on the rotation of the keypoints, we use the pre-trained pose estimation network \cite{ruiz2018fine} to compute the rotation matrix $R$.

We design an encoder-decoder model to obtain the expression deformation $\delta^{k}$. First, a Bottleneck \cite{kim2018deep} based encoder $E_{e}$ extracts expression information from the input image and compresses it into a $256$-dimensional feature vector $f_{\delta}$. Since we assume that the actual expression deformation should be related to the identity of the target face, a decoder $D_{e}$ uses the feature $f_{\delta}$ and the canonical keypoints $p_{C}^{k}$ to predict the deformation $\delta^{k}$. In this case, the estimated $\delta^{k}$ is associated only with the distribution of the canonical keypoints. The estimation of $\delta^{k}$ from the image $X$ can be represented as follows:

\begin{equation}
\small
  \begin{aligned}
  \delta_{X}^{k} = D_{e}(E_{e}(X), p_{C}^{k}). \\
  \end{aligned}
  \label{delta}
\end{equation}

With the above parameters, we can obtain the keypoints' coordinates of the source identity image $S$ and the driving motion image $D$ by:
\begin{equation}
\small
  \begin{aligned}
  p_{S}^{k} = R_{S}f_{S}(p_{C}^{k} + \delta_{S}^{k}) + t_{S}, ~
  p_{D}^{k} = R_{D}f_{D}(p_{C}^{k} + \delta_{D}^{k}) + t_{D}.
  \end{aligned}
  \label{keypoints calculation}
\end{equation}
This process is illustrated in the red box of the part (a) in Fig. \ref{fig:main}.

We follow \cite{wang2021one} to complete the subsequent dense motion estimation and target face generation. It is worth pointing out that we redesign the generator, which reconstructs results at different resolutions, denoted as $\{G_{64}, G_{128}, G\}$. Thus, we calculate the perceptual loss with the ground truths at different resolutions $\{D_{64}, D_{128}, D\}$:
\begin{equation}
\small
  \begin{aligned}
  \mathcal{L}_{P, multi-scale} = \mathcal{L}_P(D_{64}, G_{64}) + \mathcal{L}_P(D_{128}, G_{128}) + \mathcal{L}_P(D, G). \\
  \end{aligned}
  \label{multi-scale}
\end{equation}
This multi-scale generator is designed in part (c) of Fig. \ref{fig:main}. Moreover, we add the following losses to optimize the parameters of the proposed pipeline.

\textbf{Self-Supervised Representation Loss}:
During the feature extraction, we employ self-supervised representation learning to obviate the coupling of expression features with other information. Specifically, in addition to the driving image $D$, we also input a new image $D'$ by data augmentation on $D$, including rotation, scaling, and translation, to obtain the expected similar features $f_{\delta}'$ with invariant expression information. Thus, we maximize the cosine similarity between $f_{\delta}$ and $f_{\delta}'$ by minimizing:
\begin{equation}
\small
  \begin{aligned}
  \mathcal{L}_{Exp} &= 1 -\left\langle f_{\delta}, f_{\delta}'\right\rangle, \\
  \end{aligned}
  \label{cl loss}
\end{equation}
where $\langle\cdot, \cdot\rangle$ represents the cosine similarity operator. The visualization of this self-supervised representation learning is given in part (b) of Fig. \ref{fig:main}.

\textbf{Identity Latent Consistency Loss}:
In order to encourage consistent canonical keypoints, regardless of the head pose and facial attributes of the image, we use the keypoint detector to obtain the canonical keypoints ${p^{k}_{{C}_{S}}}$ and ${p^{k}_{{C}_{D}}}$ of the source image $S$ and the driving image $D$, and calculate the following loss in the canonical space:
\begin{equation}
\small
\begin{aligned}
\mathcal{L}_{C}=\left\| {p^{k}_{{C}_{S}}} - {p^{k}_{{C}_{D}}}\right\|_2.
\end{aligned}
  \label{kpc loss}
\end{equation}

\textbf{2D Landmark Loss}:
Simultaneously, to enhance the ability to extract expressions and shapes, we utilize a pre-trained landmark detector to detect 145 2D keypoints (including pupils) $L_{G}^{145}$ and $L_{D}^{145}$ from the generated image $G$ and the driving image $D$, and enforce their consistency in the 2D space with the following loss: 
\begin{equation}
\small
\begin{aligned}
\mathcal{L}_{M} = \lambda_{face} \left\| L_{G}^{120} - L_{D}^{120} \right\|_2 
+ \lambda_{mouth} \left\| L_{G}^{20} - L_{D}^{20} \right\|_2 
+ \lambda_{pupil} \left\| L_{G}^{5} - L_{D}^{5} \right\|_2,
\end{aligned}
  \label{landmark loss}
\end{equation}
where $\lambda_{face}$, $\lambda_{mouth}$ and $\lambda_{pupil}$ are hyper-parameters for different parts of the face.

\begin{figure}[t]
\begin{center}
   \includegraphics[width=0.75\linewidth]{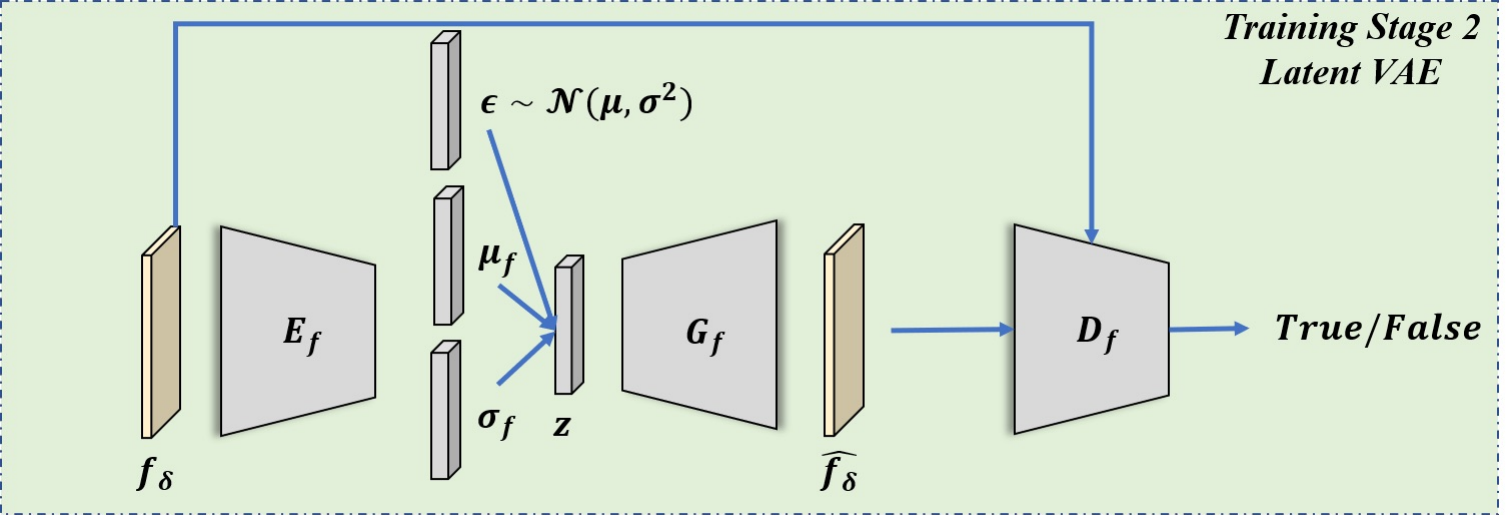}
\end{center}
   \caption{Illustration of the latent VAE training.}
\label{fig:vae}
\end{figure}

%-------------------------------------------------------------------------
\subsection{VAE Latent Space for Experssions} \label{vae}
One critical advantage of our model is the ability to actively control the expressions, which can generate arbitrary expressions for human heads under different angles without driving sources. 
We expect to reconstruct the expression latent feature $f_{\delta}$ extracted by the expression encoder $E_e$ in Sec. \ref{pipeline} and map it to a continuous distribution which is easy to sample, thus facilitating the control of the expression shape ${\delta}^k$.
Therefore, we adopt the VAE architecture, shown in Fig. \ref{fig:vae}, which can map $f_{\delta}$ to a continuous latent variable $z$ whose distribution should be close to $\mathcal{N}(0, I)$, where $I$ is the identity matrix. The expression feature encoder $E_{f}$ produces a latent Gaussian distribution $\mathcal{N}(\mu_{f}, \sigma_{f})$:
\begin{equation} \label{equ_vaeencoder}
    \begin{aligned}
        \mu_{f}, \sigma_{f} \leftarrow E_f\left( f_{\delta} \right). \\
    \end{aligned}
\end{equation}

During training, we sample the latent code from this distribution by the reparameterization trick:
\begin{equation} \label{reparameterization}
    \begin{aligned}
        z = \mu_{f} + \sigma_{f} \cdot \epsilon,
    \end{aligned}
\end{equation}
where $\epsilon \sim \mathcal{N} \left( 0, I \right)$.
The decoder $G_{f}$ produces corresponding expression feature $\hat{f_{\delta}}$ as:
\begin{equation} \label{exp feature}
    \begin{aligned}
        \hat{f_{\delta}} = G_{f} \left( z \right). \\
    \end{aligned}
\end{equation}

As a result, we optimize the following VAE loss:
\begin{equation} \label{equ_vaeloss}
    \begin{aligned}
         \mathcal{L}_{\mathrm{VAE}} =  \lambda_f \left\|f_{\delta}-G_f(\hat{f_{\delta}} \mid z)\right\|^2 
        +\lambda_{kl} \operatorname{KL}\left( \mathcal{N}(\mu_{f}, \sigma_{f}) \| \mathcal{N}(0, I)\right) 
        +\lambda_{adv} \mathcal{L}_{Adv}(f_{\delta}, \hat{f_{\delta}}). 
    \end{aligned}
\end{equation}
where $\lambda_{f}$, $\lambda_{kl}$ and $\lambda_{adv}$ are three balance hyper-parameters. During training, we observe that the balance between the KL divergence $\operatorname{KL}\left( \mathcal{N}(\mu_f, \sigma_f) \| \mathcal{N}(0, I)\right)$ and the reconstruction loss $\left\|f_{\delta}-G_f(\hat{f_{\delta}} \mid z)\right\|_2$ is prone to leading the model to generate a constant average expression. In other words, VAE will collapse before learning the distinguishable representation, mainly because the KL divergence converges faster than the reconstruction loss, which indicates that the encoder is no longer functional \cite{zhang2019learning}. We introduce the adversarial loss \cite{tang2022explicitly} $\mathcal{L}_{Adv} (f_{\delta}, \hat{f_{\delta}})$ to solve this problem. Experimental evidence shows that this can guarantee the diversity in the distribution of the expression features.

%------------------------------------------------------------------------
\subsection{Training}
The full model MMFA is trained with the same-identity reconstruction in   a self-supervised manner. Specifically, we randomly select a pair of images in the same video sequence and set one frame as the source image $S$, while another is used as the driving image $D$ and also as the reconstruction ground truth.
The training loss of MMFA is:
\begin{equation}
\small
  \begin{aligned}
  \mathcal{L}_{total} = \mathcal{L}_{P, multi-scale} + \mathcal{L}_{\mathrm{GAN}}(D, G) +\mathcal{L}_E(\{p_{D}^{k}\}) + \\
  \mathcal{L}_L(\{p_{D}^{k}\}) + \mathcal{L}_{\Delta}(\{\delta_{D}^{k}\}) + \mathcal{L}_{Exp} + \mathcal{L}_{C} + \mathcal{L}_{M},
  \end{aligned}
  \label{All_loss}
\end{equation}
where $G$ is the final output image of the whole network.
In the training of our model, the multi-scale perceptual loss $\mathcal{L}_{P, multi-scale}$ and GAN loss $\mathcal{L}_{\mathrm{GAN}}$ mainly act as the reconstruction loss for the quality of generated result $G$. The equivariance loss $\mathcal{L}_E$ is used as the constraint to ensure the consistency of keypoints \cite{siarohin2019first}. $\mathcal{L}_{L}$ and $\mathcal{L}_{\Delta}$ are two effective prior losses for 3D keypoints proposed in \cite{wang2021one}. Besides, $\mathcal{L}_{Exp}$, $\mathcal{L}_{C}$ and $\mathcal{L}_{M}$ have been described in Sec. \ref{pipeline}.

Note that the training of the latent VAE described in Sec. \ref{vae} is independent of the training of MMFA. After the training of MMFA and the latent VAE, we can use the VAE to manipulate the expression generation by MMFA.

\begin{figure}[tp]
\begin{center}
   \includegraphics[width=0.9\linewidth]{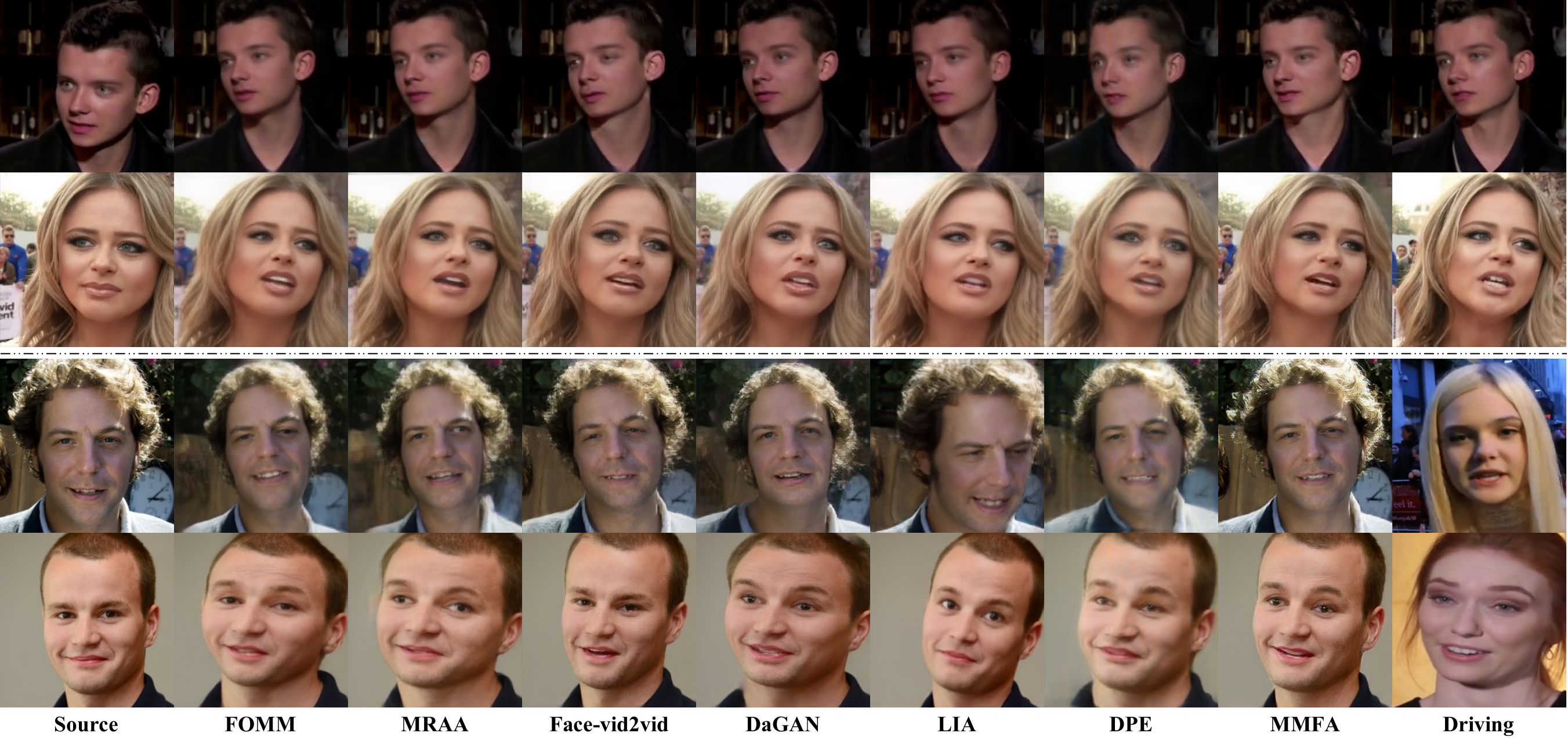}
\end{center}
   \caption{Visual comparisons with state-of-the-art methods.}
\label{fig:result1}
\end{figure}

\begin{table}[t]
 \caption{Quantitative comparison of face animation on VoxCeleb \cite{nagrani2017voxceleb}.}
\centering
\resizebox{\textwidth}{!}{%
\begin{tabular}{llllllllllll}
    \toprule
\multirow{2}{*}{Method} & \multicolumn{7}{c}{Same-identity}  & \multicolumn{4}{c}{Cross-identity} \\ \cmidrule(r){2-8}   \cmidrule(r){9-12}
  & LPIPS $\downarrow$ & CSIM $\uparrow$  & PSNR $\uparrow$  & APD $\downarrow$ & AKD $\downarrow$ & AED $\downarrow$ & FID $\downarrow$ & CSIM $\uparrow$ & APD $\downarrow$   & AED $\downarrow$ & FID $\downarrow$   \\ \midrule
FOMM \cite{siarohin2019first}  
& 0.108  & 0.965   & 23.340    & \textbf{0.010}   & 1.300 & 0.0223 & 22.755    & 0.903   & 0.027   & 0.0768   & 105.216 \\
MRAA \cite{siarohin2021motion} 
& 0.101  & 0.970   & \textbf{24.641}   & 0.011   & 1.209 &0.0218 & 21.397    & 0.901  &  0.026   & 0.0766 & 87.276  \\
Face-vid2vid \cite{wang2021one}
& 0.107   & 0.972  & 22.620  & 0.014   & 1.486 & 0.0245 & 20.092    & 0.927  & 0.033  & 0.0764 & 86.054 \\
DaGAN \cite{hong2022depth}    
& \textbf{0.094}  & \textbf{0.974}  &24.168  & 0.011   & \textbf{1.198} & \textbf{0.0210} & 14.463    & 0.907       & \textbf{0.023}  &0.0770 &  86.012 \\
LIA \cite{wang2022latent}      
& 0.106 & 0.967    & 23.670   &  0.012  & 1.391 & 0.0226 & 23.992   & \textbf{0.936}     & 0.141    & 0.0774  &  83.065 \\
DPE \cite{pang2023dpe}  
&  0.127 & 0.967  & 23.481 & 0.019   & 1.739  &0.0243 &  29.620   & 0.887   & 0.043     &  0.0763 & 128.904    \\ 
\midrule
\textbf{MMFA}
& 0.106     & \textbf{0.974}   & 22.898  & 0.017  & 1.476 & 0.0241 & \textbf{13.265}  & 0.925    & 0.042 & \textbf{0.0762} & \textbf{77.445}  \\ 
    \bottomrule
    \end{tabular}%
  }
\label{table:result}
\end{table}

\section{Experiments}

%-------------------------------------------------------------------------
\subsection{Settings}
% \textbf{Implementation Details.} The face animation training of MMFA is conducted on the training set of VoxCeleb \cite{nagrani2017voxceleb}, by selecting both the source image and the driving image randomly in the same video for self-supervised learning. We use Adam \cite{kingma2014adam} with $\beta_1 = 0.5$ and $\beta_2 = 0.9$ to optimize our model on 4 24GB NVIDIA 3090 GPUs, with the learning rate $\eta = 2 \times 10^{-4}$. Besides, the training of the latent VAE is also performed on VoxCeleb similarly.

\textbf{Benchmark Datasets.}
The training of MMFA is conducted on the training set of VoxCeleb \cite{nagrani2017voxceleb}.
% For evaluation, we collect images of different identities from CelebA \cite{liu2018large} and FFHQ \cite{karras2019style}, as well as videos from the VoxCeleb test set, to form 80 image-video pairs. These images and videos are unseen during the training process. For same-identity reconstruction, we use the first frame as the source image and the remaining frames as the driving images. For cross-identity reenactment, we use video frames to drive the source image in each image-video pair. Therefore, we can evaluate the 25K synthesized images obtained by each method.
For evaluation, we collect images of different identities from CelebA \cite{liu2018large} and FFHQ \cite{karras2019style}, as well as videos from the VoxCeleb test set, to form 80 image-video pairs.
% These images and videos are unseen during the training process.
For same-identity reconstruction, we use the first frame as the source image and the remaining frames as the driving images. For cross-identity reenactment, we use video frames to drive the source image in each image-video pair. Therefore, we can evaluate the 25K synthesized images obtained by each method.

\textbf{Evaluation Metrics.} 
We adopt multiple metrics to evaluate the quality of animation. Learning Perception Image Patch Similarity (LPIPS) \cite{zhang2018perceptual}, Peak Signal to Noise Ratio (PSNR), and Structural Similarity Index (SSIM) \cite{ZhouWang2004ImageQA} are used to measure reconstruction accuracy. Fréchet Inception Distance (FID) \cite{MartinHeusel2017GANsTB} measures output visual quality. We also use cosine similarity (CSIM) \cite{zeng2022fnevr} of facial identity embeddings and AED \cite{siarohin2019first} to evaluate the ability of identity retention. For testing motion transfer quality, we employ APD \cite{ren2021pirenderer} and AKD \cite{siarohin2019first} to calculate posture and average landmark distance.

%-------------------------------------------------------------------------
\subsection{One-Shot Face Animation}
We mainly select several state-of-the-art face animation methods based on unsupervised keypoint positioning as baselines: FOMM \cite{siarohin2019first}, Face-vid2vid \cite{wang2021one}, DaGAN \cite{hong2022depth}, and MRAA \cite{siarohin2021motion}. In addition, we also select several methods which animate faces in a latent space for comparison: LIA \cite{wang2022latent} and DPE \cite{pang2023dpe}. Except Face-vid2vid, we use the officially released pre-trained models. For the experiments, since \cite{wang2021one} does not release its Face-vid2vid model, we use the model implemented at this site \footnote{\url{https://github.com/zhanglonghao1992/One-Shot_Free-View_Neural_Talking_Head_Synthesis}}.

\textbf{Qualitative Analysis.}
The visual results of the same-identity and cross-identity are shown in the first three rows and the last three rows of Fig. \ref{fig:result1}, respectively. MMFA can clearly achieve better image clarity and quality than other methods. For same-identity reconstruction, FOMM, MRAA, and DaGAN based on unsupervised keypoint positioning can achieve approximately good driving performance, and DPE may exhibit significant blur. Our MMFA also performs better on the mouth (such as the teeth) than other models. For cross-identity reenactment, FOMM, MRAA, and DaGAN relying solely on 2D keypoint positioning suffer from significant identity loss when the facial shape difference between the driving face and the source face is notably large. Face-vid2vid and MMFA based on 3D keypoint positioning and LIA based on latent space mapping are able to preserve identity information to the maximum extent with most natural facial expressions.

\textbf{Quantitative Analysis.}
The quantitative results of the two tasks are shown in Table \ref{table:result}. The FID of MMFA is the best, indicating that the faces we generate are closer to the original images than other methods. For same-identity reconstruction, DaGAN and MMFA based on unsupervised keypoint positioning have better performances overall. In the cross-identity reenactment, FOMM, MRAA, and DaGAN show lower ability to maintain identity in terms of CSIM, suggesting that these methods have obvious face deformation when there is a large shape difference and cannot well keep the identity information for face recognition (also see Fig. \ref{fig:result1}). While LIA demonstrates superior performance on the CSIM metric compared to other models, its performance is notably worse in terms of APD in cross-identity testing. Meanwhile, our best AED and suboptimal CSIM indicate that MMFA can operate stably in wild environments. 

\begin{figure}[t]
\begin{center}
   \includegraphics[width=0.9\linewidth]{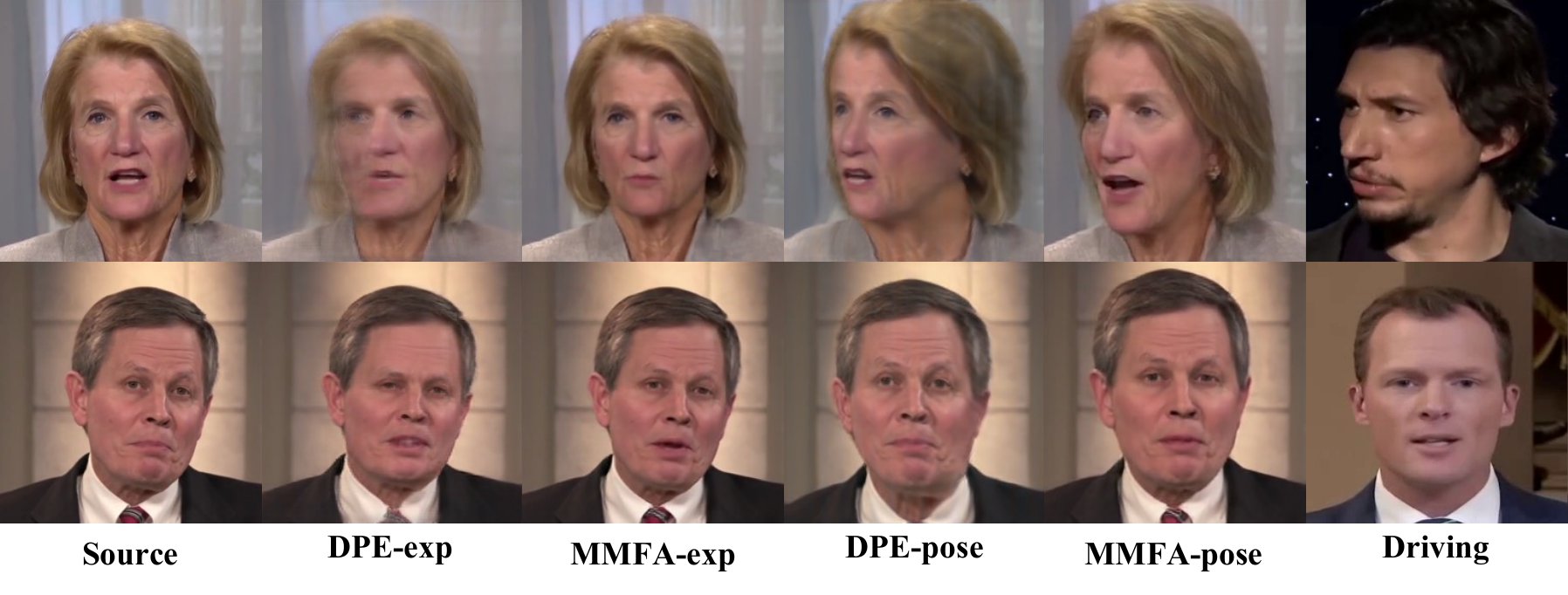}
\end{center}
   \caption{Comparison with DPE \cite{pang2023dpe} for independent pose and expression editing.}
\label{fig:dpe}
\end{figure}

\begin{figure}[t]
\begin{center}
   \includegraphics[width=0.9\linewidth]{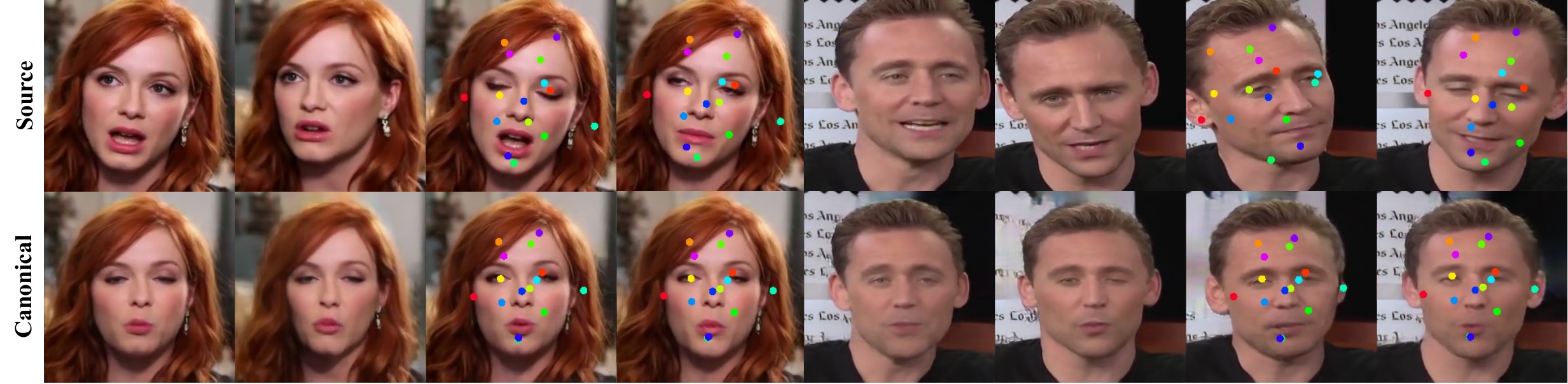}
\end{center}
   \caption{Visual comparisons with state-of-the-art methods.}
   
\label{fig:Canonical}
\end{figure}

\begin{figure}[t]
\begin{center}
   \includegraphics[width=0.9\linewidth]{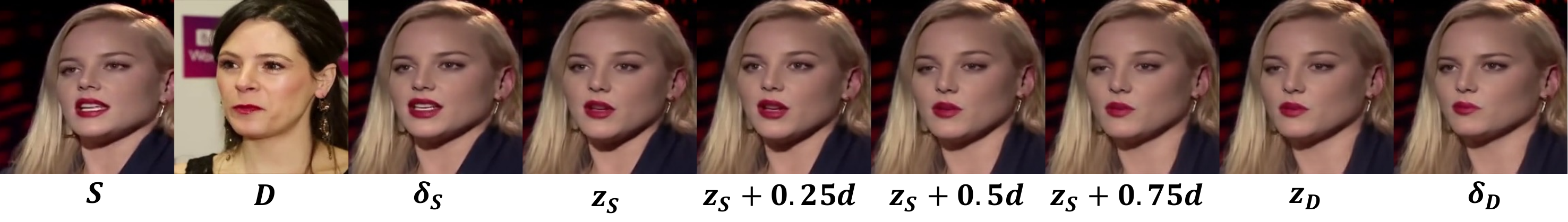}
\end{center}
   \caption{Results of interpolation in the VAE latent space. The last seven columns represent the face of $S$ with the pose of $D$, the generated results with the VAE interpolation ($d = z_D - z_S$), and the face of $S$ with the pose and expression of $D$, respectively.}
\label{fig:vae_exp}
\end{figure}

\begin{figure}[t]
\begin{center}
   \includegraphics[width=0.8\linewidth]{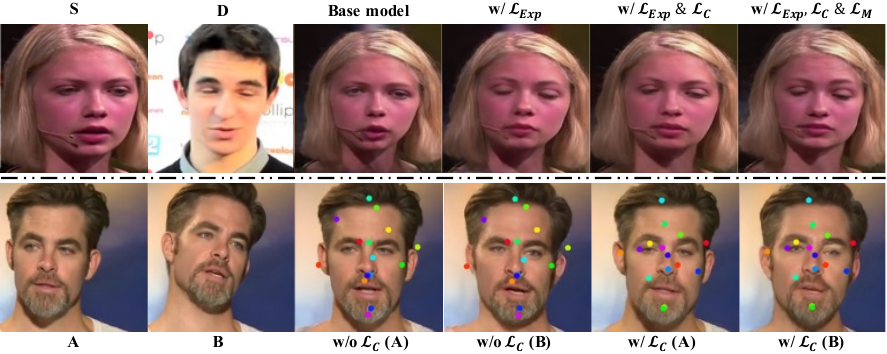}
\end{center}
   \caption{Ablation study of MMFA.}
\label{fig:ablation}
\end{figure}

%-------------------------------------------------------------------------
\subsection{Motion Attribute Editing}
A few one-shot face animation methods \cite{ren2021pirenderer, wang2022latent, pang2023dpe} have been developed to independently edit expressions or poses, making them appropriate for general video portrait editing. In this study, we compare our MMFA to the state-of-the-art open-source model DPE \cite{pang2023dpe}, which performs independent editing in the latent space. However, as shown in Fig. \ref{fig:dpe}, when attempting to transfer larger poses or expressions, DPE suffers from significant identity loss and lacks facial details. Besides, DPE is unable to achieve the explicit control and deeper decoupling of motion attributes as our MMFA, which allows actively specifying the pose, scale, and position of the face on the image plane. Furthermore, another crucial advantage of our MMFA is that the keypoints provide explicit spatial information, avoiding the influence on the background, while DPE produces significant distortions to areas outside the face, such as the tie of the man in Fig. \ref{fig:dpe}.
% A few one-shot face animation methods have been developed to independently edit expressions or poses, making them appropriate for general video portrait editing. Some of them are based on prior information from 3DMM \cite{yin2022styleheat, ren2021pirenderer, yang2022face2face}, while others rely on self-supervised disentanglement in a latent space \cite{pang2023dpe, wang2022latent}. In this study, we compare our MMFA to the state-of-the-art open-source model DPE \cite{pang2023dpe}, which performs independent editing in the latent space by adding motion code from the driving image to the source image's latent code. However, as shown in Fig. \ref{fig:dpe}, when attempting to transfer larger poses or expressions, DPE suffers from significant identity loss and lacks facial details. Furthermore, DPE is unable to achieve the same explicit control and deeper decoupling of motion attributes as our MMFA. MMFA can not only transfer the motion according to the driving image, but also allow for actively specifying the pose, scale, and position of the face on the image plane.
 
We visualize MMFA's canonical faces in Fig. \ref{fig:Canonical}. Although there are significant differences in pose and expression between the source images, the semantic images represented by the canonical keypoints can capture almost identical identities, indicating that MMFA can retain important identity information. And the images of cross-identities have almost the same neutral posture, expression, and size, which confirms the feasibility of our cross-identity motion attribute editing.
%-------------------------------------------------------------------------
\subsection{Expression Interpolation with the Latent VAE}
Fig. \ref{fig:vae_exp} shows the generated results by our trained latent VAE for two pairs of source images $S$ and driving images $D$ after the interpolation of features $z$ in the latent space, where $D$ provides the affine transformation parameters and target expression, and $S$ provides the face identity. It can be seen that, in both same-identity reconstruction and cross-identity reenactment, our latent space has an effective continuous control ability for expression changes. This expression control can be applied to various portrait editing scenarios.

%-------------------------------------------------------------------------
\subsection{Ablation Study}
We further conduct an ablation study on $\mathcal{L}_{Exp}$, $\mathcal{L}_{C}$ and $\mathcal{L}_{M}$ to evaluate their effectiveness, and the visualization results are shown in Fig. \ref{fig:ablation}. 
% We take the MMFA trained without the three loss functions as the base model, and construct three comparison models, each of which introduces one loss respectively.  
As shown in the first row of Fig. \ref{fig:ablation}, where the base model denotes the MMFA trained without the above three losses, we can observe that $\mathcal{L}_{Exp}$ is effective to ensure the extraction of accurate expression features, and $\mathcal{L}_{M}$ can generate natural expressions and improve model's ability to maintain facial shape. 
% can improve the quality of the generated results. 
To intuitively present the effectiveness of $\mathcal{L}_{C}$, we select two examples with large pose variations and visualize the predicted canonical faces in the second row of Fig. \ref{fig:ablation}. It shows that $\mathcal{L}_{C}$ can alleviate the distortions of canonical facial details and helps to predict more consistent canonical keypoints for the same identity regardless of the pose changes, demonstrating that $\mathcal{L}_{C}$ can improve the robustness to large pose variations.

% \lxh{Although $\mathcal{L}_{C}$ slightly weakens the ability of expression parsing, it is robust to large pose changes in the detection of canonical keypoints.}
% It shows that  on the canonical keypoints, making the keypoint detector robust to large pose variations.
%------------------------------------------------------------------------
\section{Conclusion}
In this paper, we propose a new unsupervised pipeline MMFA for computing facial keypoints to drive facial animation with a high degree of expression manipulation and detail transfer, achieved through variational autoencoder training. Our method can disentangle attributes such as expression, pose, and identity, enabling their independent control, and manipulate facial features to elicit intended expressions and motions. Extensive experiments validate that MMFA not only provides excellent expression control ability but also has more detailed expression transfer and generation authenticity.

{\small
\bibliographystyle{unsrtnat}
\setcitestyle{numbers,open={[},close={]}} 
\bibliography{reference}
}

% \section*{Supplementary Material}

\clearpage
\appendix
\section*{Supplementary Material}
In this supplement, we first introduce the additional network architectures and loss functions (Sec.~\ref{network_and_loss}). Then, we further analyze the additional experimental details (Sec.~\ref{metric_and_optim}). After that, we show more visual results of the same-identity-driven, cross-identity-driven, motion attribute editing and expression interpolation with the latent VAE, as well as quantitative ablation study (Sec. \ref{results}). Finally, we discuss the social impact and limitations (Sec. \ref{social} and Sec. \ref{limit}).

\section{Additional Network Architectures and Training Losses} \label{network_and_loss}
In this section, we present the architecture of MMFA and discuss the losses in detail.

\subsection{Network Architectures}
The architectures of the keypoint detector, affine transformation, and dense motion estimator are similar to \cite{wang2021one}. We mainly introduce the new network expression encoder-decoder and multi-scale generator adopted by MMFA.

\textbf{Expression Encoder-Decoder.}
The expression encoder-decoder extracts expression feature $f_{\delta}$ and predicts keypoint deformation $\delta^{k}$. Specifically, the encoder first uses a convolutional layer with ReLU activation and a maxpooling layer to extract image features. Subsequently, we employ ResBottleneck blocks \cite{kim2018deep} to downsize the image features and obtain a $16 \times 4 \times 4$ feature map. Finally, we add an additional linear layer to map the reshaped feature map to a $256$-dimensional expression feature $f_{\delta}$.

Given that the expression feature $f_{\delta}$ is a linear vector, we directly concatenate it with the canonical keypoints $p_{C}^{k}$, which provides the spatial information for deformation prediction. The decoder is composed of several fully connected layers with BatchNorm layer \cite{ioffe2015batch} and ReLU activation, which are used to process the concatenated input. The final layer of the decoder is a simple fully connected layer without activation since the expression deformation can be negative. 

\textbf{Multi-Scale Generator.}
To improve the generation of the output image, we design a multi-scale generator with output faces at different resolutions. Firstly, we reshape the warped $32 \times 16 \times 64 \times 64$ 3D feature map and use a convolutional layer to process it to $256 \times 64 \times 64$ 2D feature map.

Then, we use a ResBlock \cite{kim2018deep} to extract the image feature. In addition, we also employ a convolutional layer to change the feature channel and obtain an output image at this resolution. Subsequently, we concatenate the image feature and the output image. To increase the concatenated feature's resolution and reduce its channel, we adopt an UpBlock with an up-sampling layer and a convolutional layer. The above process is repeated 3 times, and we obtain the features and output images $\{G_{64}, G_{128}, G\}$ at resolution $64 \times 64$, $128 \times 128$ and $256 \times 256$.

\subsection{Losses}
In the main paper, we already discuss $\mathcal{L}_{Exp}$, $\mathcal{L}_{C}$ and $\mathcal{L}_{M}$. Thus, we elaborate on the remaining losses in this section.

\textbf{Multi-Scale Perceptual Loss $\mathcal{L}_{P, multi-scale}$.} Since we have explained the formulation of the multi-scale loss for generated images $\{G_{64}, G_{128}, G\}$ and ground truths $\{D_{64}, D_{128}, D\}$ in the main paper, here, we show the computation details of the VGG perceptual loss \cite{johnson2016perceptual}, which calculates the difference between the generated result and the ground truth by extracting the semantic features of the image through a pre-trained neural network. We use both VGG-19 \cite{simonyan2015very} and VGG-Face \cite{parkhi2015deep}
to compute the high-level features. For VGG-Face, we only obtain the loss with generated image $G$ at $256 \times 256$ resolution. For VGG-19, we perform the loss computation with a down-sampling pyramid for the output $\{G_{64}, G_{128}, G\}$ and the ground truth $\{D_{64}, D_{128}, D\}$, respectively.

\textbf{GAN Loss $\mathcal{L}_{\mathrm{GAN}}$.}
We employ a discriminator which outputs the intermediate features and final results to distinguish the virtual face image from the ground truth. With the intermediate features, we compute the feature matching loss \cite{wang2018high} to constrain the generation of images in multi-scale features. With the discriminator's final output, we use the least square loss \cite{mao2017least} to train an adversarial generative network.

\textbf{Equivariance Loss $\mathcal{L}_E$.}
Following \cite{siarohin2019first}, we also adopt the equivariance loss to ensure that all keypoints have fixed semantics. When a known geometric transformation $\mathcal{T}$ is applied on an image $X$ with keypoints $p_{X}^{k}$, the new estimated keypoints $p_{\mathcal{T}(X)}^{k}$ for transformed image $\mathcal{T}(X)$ should indicate the same region of the face:
\begin{equation}
\mathcal{L}_{E}=\sum_{k=1}^{K}\left\lVert p_{X}^{k}-\mathcal{T}^{-1}\!\left(p_{\mathcal{T}(X)}^{k}\right)\right\rVert_{1}.
\label{equivariance_loss}
\end{equation}
where $\mathcal{T}^{-1}$ is the inverse transform of $\mathcal{T}$. Note that we project the 3D keypoints to 2D for loss calculation and when we compute $p_{\mathcal{T}(X)}^{k}$ with our pipeline, the canonical keypoints of $\mathcal{T}(X)$ should be estimated since the identity of the person is changed after the geometric transformation.

\textbf{Keypoint Prior Loss $\mathcal{L}_{L}$.}
This loss is used to constrain the keypoints' 
locations with a threshold $D_{t}$ for the distance between each pair of them and a threshold $z_{t}$ for the mean depth. We use the same setting of threshold as in \cite{wang2021one}.

\textbf{Deformation Prior Loss $\mathcal{L}_{\Delta}$.}
It is crucial to limit the magnitude of the keypoint deformation $\delta^{k}$ to prevent it from affecting the identity of the face represented by the canonical keypoints. Specifically, we directly compute its $\mathcal{L}_1$ norm and set the norm as a loss to constrain the keypoint deformation's value.

\section{Additional Experiment Details} \label{metric_and_optim}
\subsection{Metrics}
We adopt various metrics to evaluate the quality of the animations. For same-identity driving, we use LPIPS \cite{zhang2018perceptual} and PSNR to measure the fidelity of video reconstruction, FID \cite{MartinHeusel2017GANsTB}  to assess the distribution consistency between the reconstructed and original videos, CSIM \cite{zeng2022fnevr} and AED \cite{siarohin2019first} to evaluate the similarity of facial identity features between the reconstructed and original videos, and APD \cite{ren2021pirenderer} and AKD \cite{siarohin2019first}  to measure the motion transfer quality of the reconstructed video. For cross-identity driving, we use CSIM and AED to measure the similarity of facial identity features between the reenacted video and the source image, APD to assess the facial pose transfer capability between the reproduced video and the driving video, and FID to determine the closeness of the distribution between the reproduced video and the source image.
We describe the implementation details for these metrics:

\textbf{LPIPS} measures the difference between images in terms of human visual perception by extracting image features using pre-trained deep convolutional neural networks (such as VGG or AlexNet) and calculating the distance between these features.

\textbf{PSNR} measures the degree of image distortion by calculating the mean squared error (MSE) between the original images and the reconstructed images.

\textbf{FID} measures the statistical difference between real images and generated images by calculating the Fréchet distance of features extracted from a pre-trained Inception network.

\textbf{AED} measures the degree of identity preservation by calculating the Average Euclidean Distance between the identity features extracted from real face images and generated face images using a pre-trained face recognition network \cite{baltruvsaitis2016openface}.

\textbf{APD} measures the pose transferability by calculating the $\mathcal{L}1$ distance between the pose parameters of real face images and generated face images, which are extracted from Deep3D \cite{deep3d2020} officially released by Microsoft.

\textbf{AKD} measures the motion preservation capability by calculating the average keypoint distance between real face images and generated face images using a pre-trained facial landmark detector \cite{bulat2017far}.

\subsection{Optimization}
The face animation training of MMFA is conducted on the training set of VoxCeleb \cite{nagrani2017voxceleb}, by selecting both the source image and the driving image randomly in the same video for self-supervised learning. We use Adam \cite{kingma2014adam} with $\beta_1 = 0.5$ and $\beta_2 = 0.9$ to optimize our model on 4 24GB NVIDIA 3090 GPUs, with the learning rate $\eta = 5 \times 10^{-5}$. Besides, the training of the latent VAE is also performed on VoxCeleb similarly.

\section{Additional Experimental Results} \label{results}

\subsection{One-Shot Face Animation}

\begin{figure}[t]
\begin{center}
   \includegraphics[width=\linewidth]{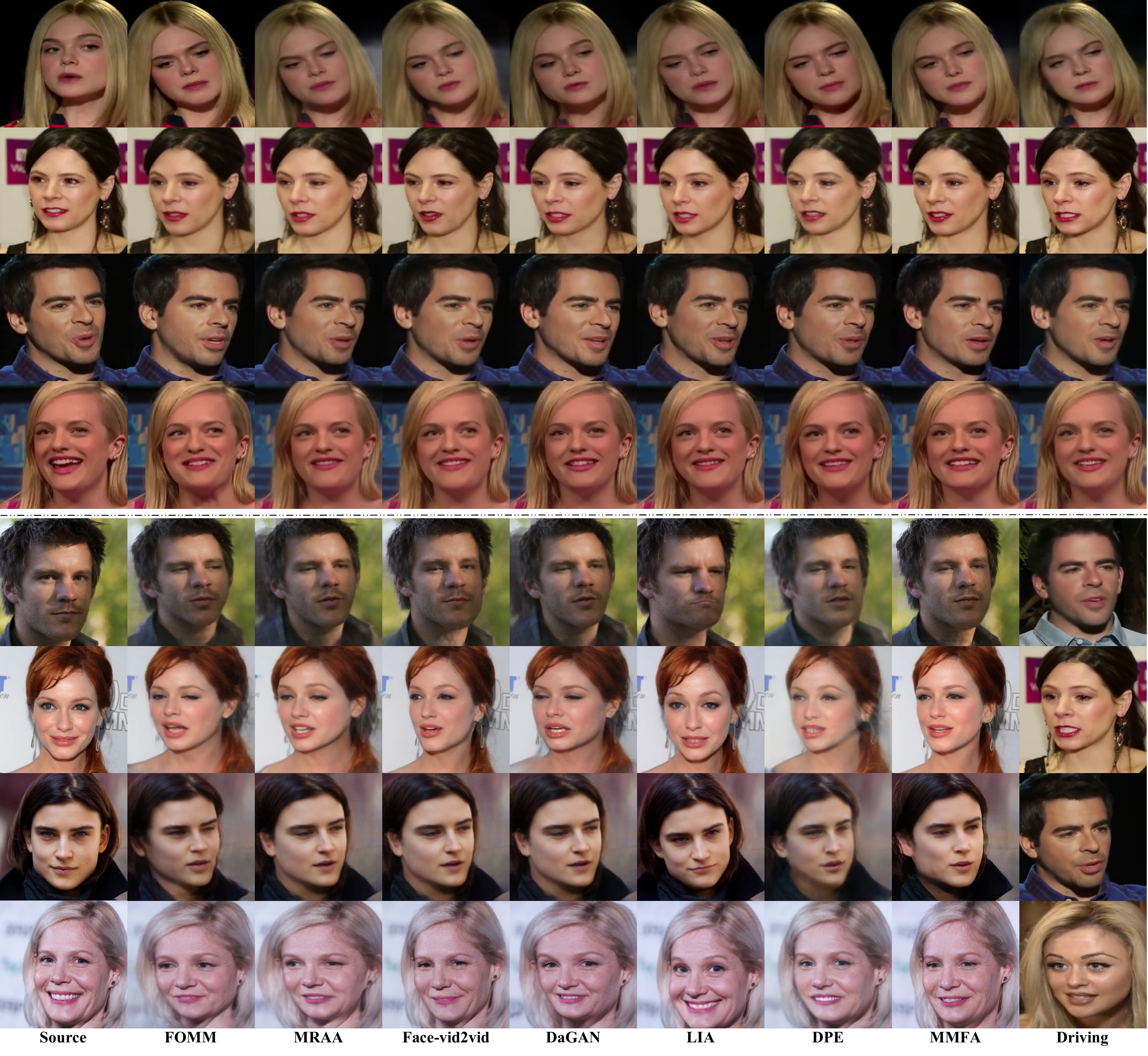}
\end{center}
   \caption{Visualization comparison of our MMFA with other state-of-the-art methods.}
   
\label{fig:sup1}
\end{figure}

As shown in Fig. \ref{fig:sup1}, we present a comparison of visual results between our MMFA and other state-of-the-art methods. The first four rows show results driven by the same identity. It can be observed that MMFA is capable of generating clearer images and more natural facial expressions. However, it performs less well in eyeball tracking compared to methods driven by 2D key points, such as FOMM \cite{siarohin2019first} and MRAA \cite{siarohin2021motion}, as demonstrated in the fourth row where these methods accurately track the eyeball position. 
The remaining four rows display cross-identity face animation results. MMFA achieves advantages in identity preservation, expression transfer, and image clarity. As shown in the seventh and eighth rows, FOMM, MRAA, and DaGAN \cite{hong2022depth} exhibit noticeable shape distortions, leading to identity information loss. DPE \cite{pang2023dpe} generates more blurred results, LIA \cite{wang2022latent} fails to properly handle relative pose and expression, and Face-vid2vid's \cite{wang2021one} expression driving is less accurate. These visualization results further complement the qualitative and quantitative analysis presented in Sec. 4.2 of the manuscript.

\subsection{Motion Attribute Editing}

\begin{figure}[t]
\begin{center}
   \includegraphics[width=\linewidth]{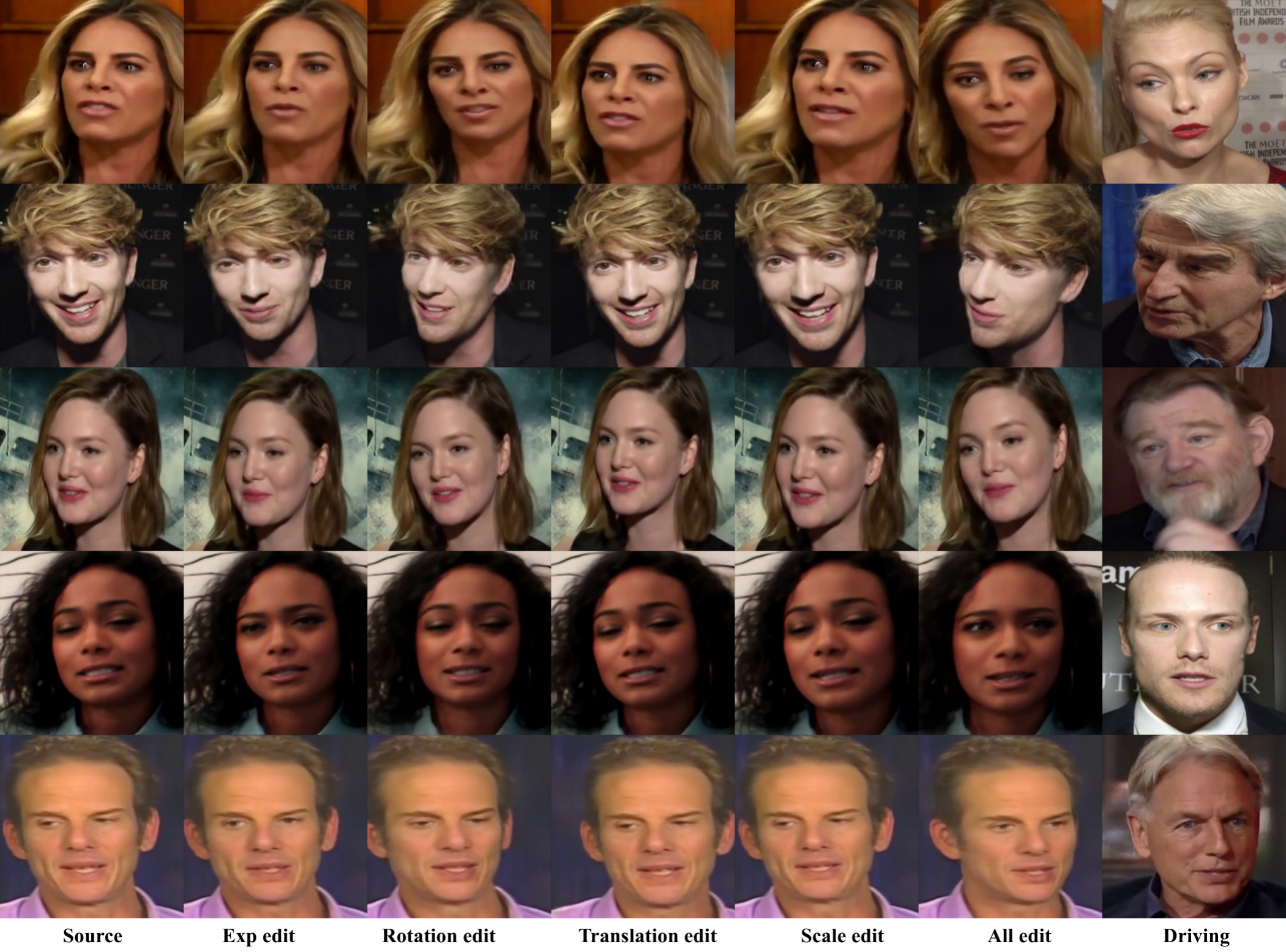}
\end{center}
   \caption{More visualization results of motion attribute editing.}
   
\label{fig:edit}
\end{figure}

\begin{figure}[t]
\begin{center}
   \includegraphics[width=\linewidth]{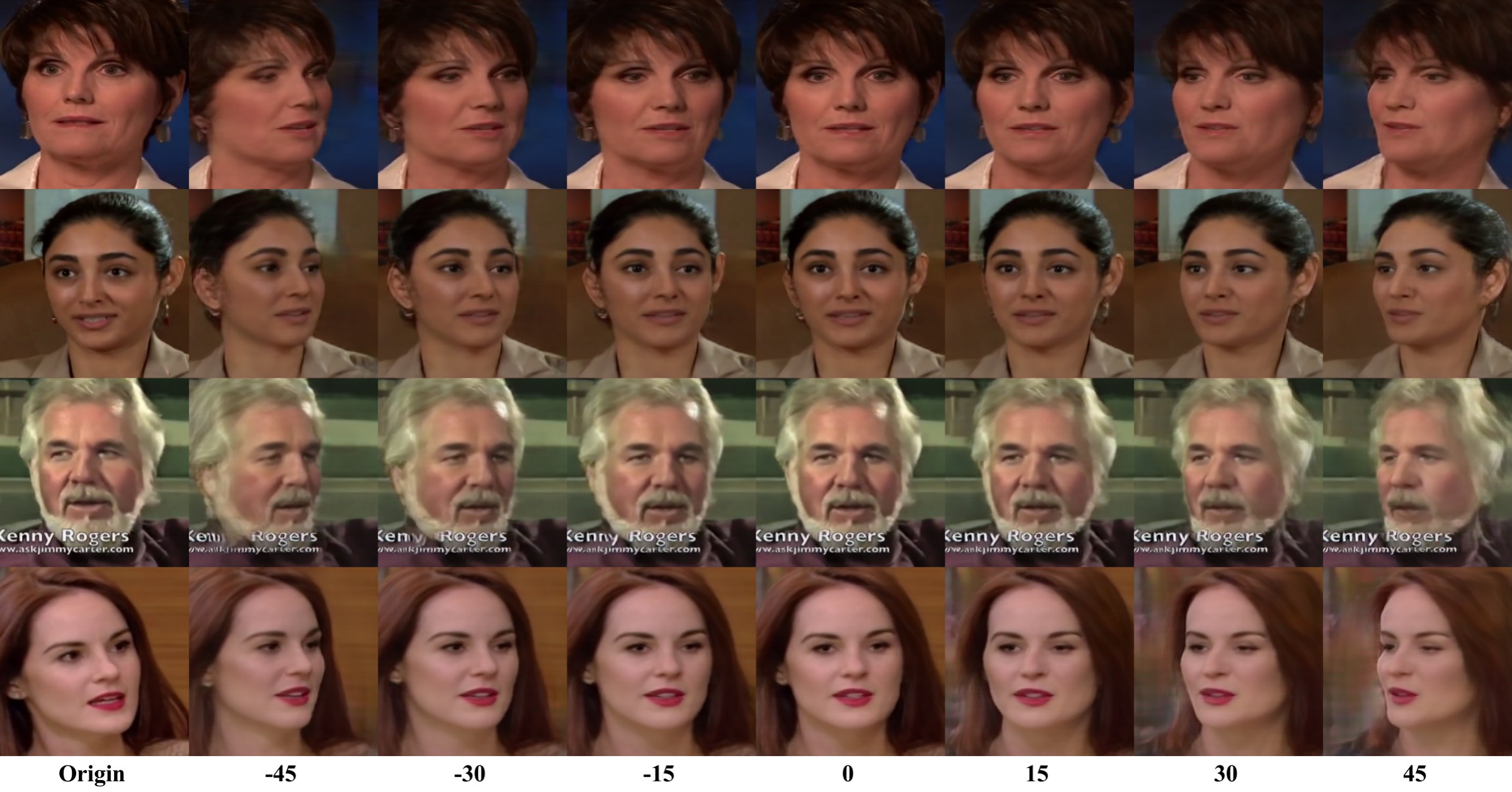}
\end{center}
   \caption{Visualization results of active editing of yaw angle.}
   
\label{fig:yaw}
\end{figure}

\begin{figure}[t]
\begin{center}
   \includegraphics[width=\linewidth]{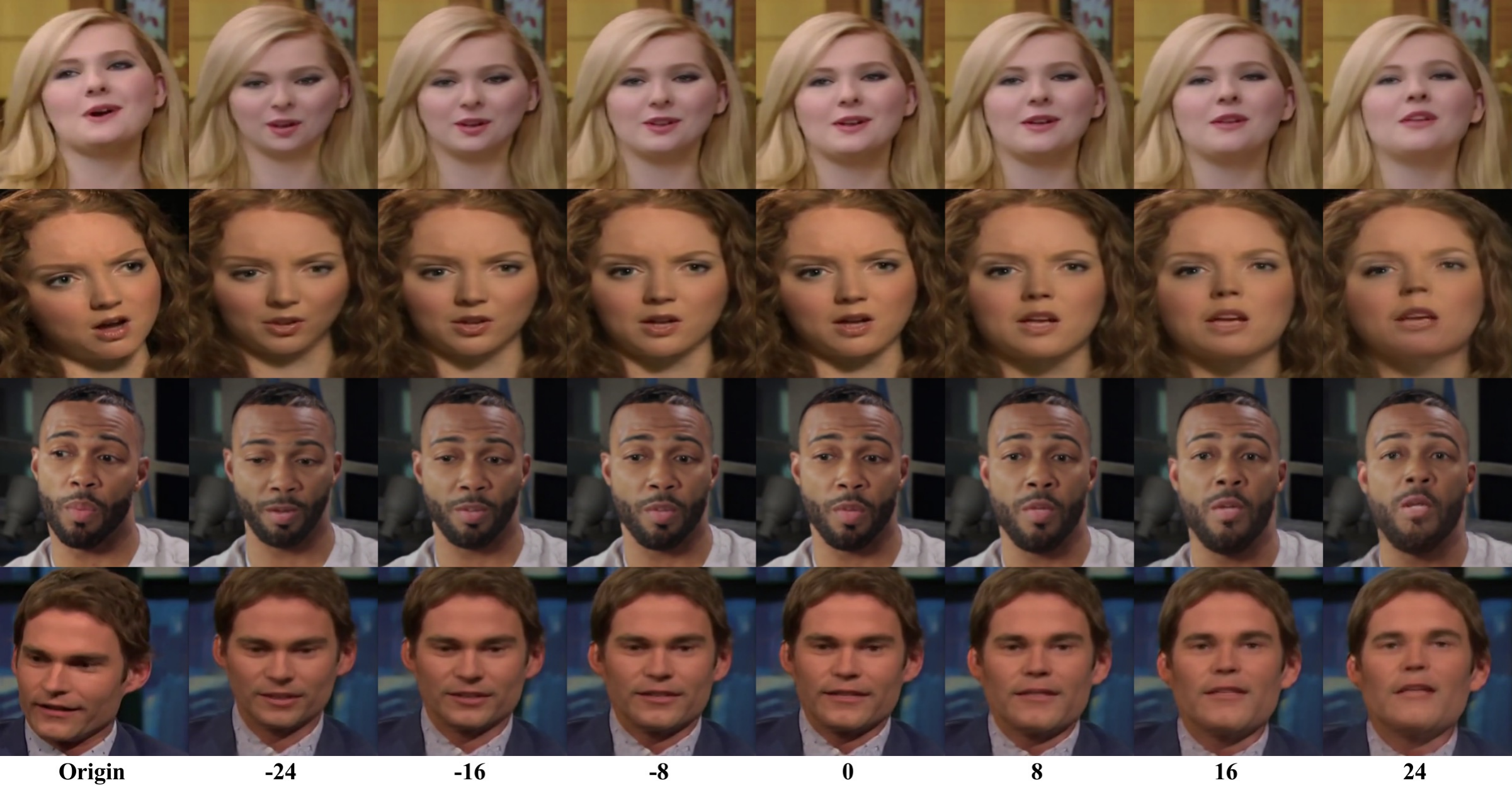}
\end{center}
   \caption{Visualization results of active editing of pitch angle.}
   
\label{fig:pitch}
\end{figure}

\begin{figure}[t]
\begin{center}
   \includegraphics[width=\linewidth]{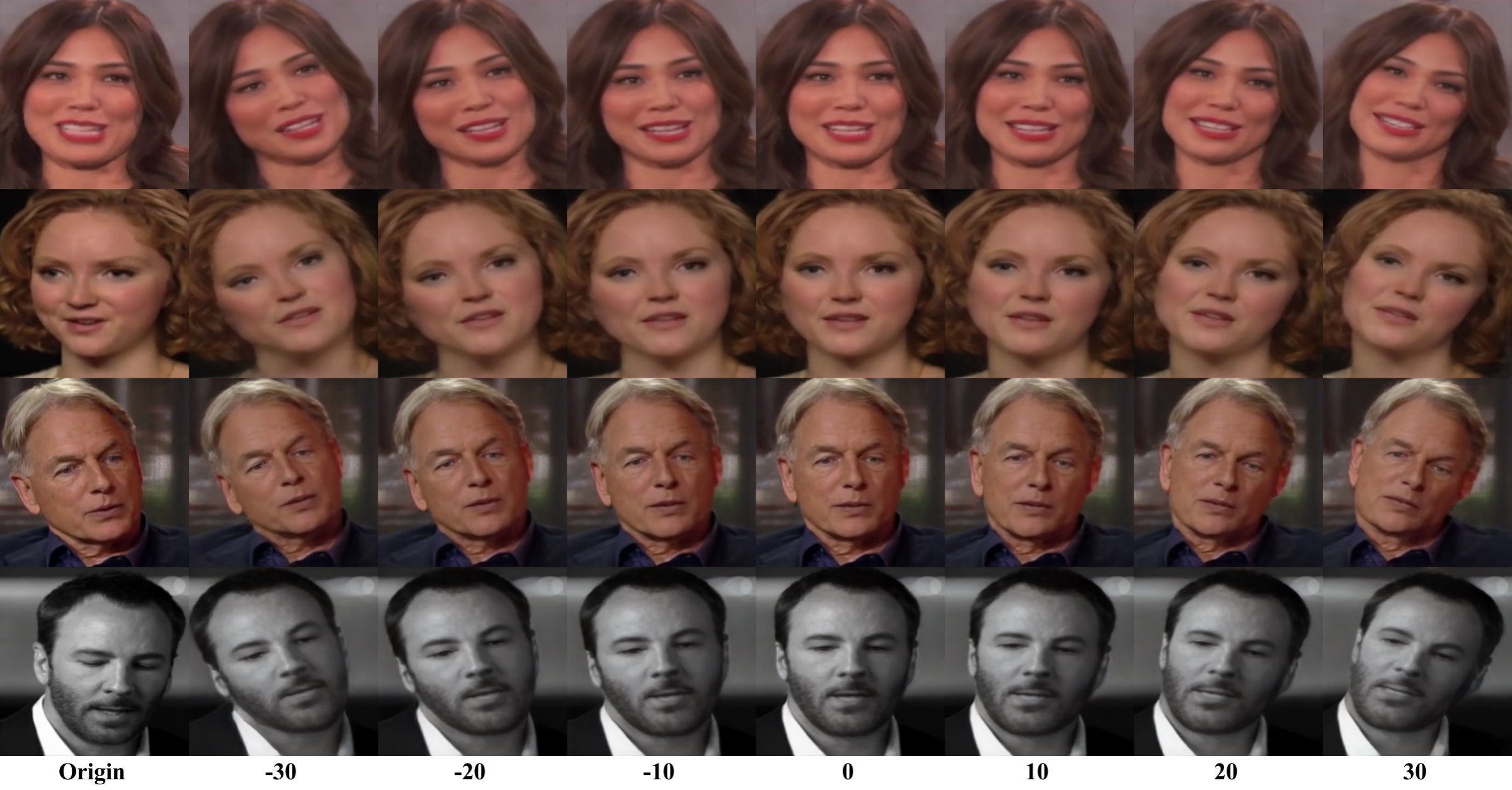}
\end{center}
   \caption{Visualization results of active editing of roll angle.}
   
\label{fig:roll}
\end{figure}

In this section, we present more visualization results to demonstrate the effectiveness of our motion attribute editing. These examples show how our method manipulates various motion attributes, such as facial expressions, head poses, and spatial translations while preserving the subject's identity. We transfer attributes like expressions, rotations, translations, and scaling from the driving image to the source image separately. Fig. \ref{fig:edit} indicates that MMFA can perform motion attribute editing individually or as a whole while maintaining image generation quality and identity consistency.
Additionally, we also demonstrate the active editing effects for head pose attributes. Figs. \ref{fig:yaw}, \ref{fig:pitch}, and \ref{fig:roll} visualize the results of different yaw angles, pitch angles, and roll angles editing, respectively. This verifies that MMFA has the potential to generate animations of the subject's head moving smoothly and naturally in different directions.
These additional visualization results further confirm the versatility and effectiveness of our motion attribute editing method, which can preserve identity information to the greatest extent and allow the generation of high-quality and realistic face animation across various scenarios.

\subsection{Expression Interpolation with the Latent VAE}
\begin{figure}[t]
\begin{center}
   \includegraphics[width=\linewidth]{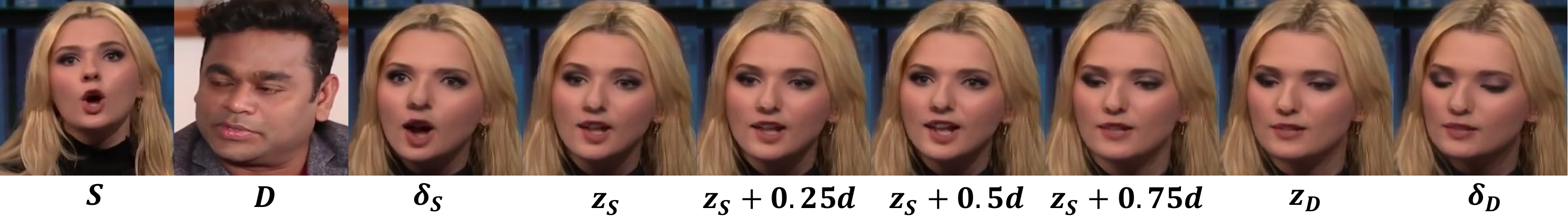}
\end{center}
   \caption{Results of interpolation in the VAE latent space. The last seven columns represent the face of $S$ with the pose of $D$, the generated results with the VAE interpolation ($d = z_D - z_S$), and the face of $S$ with the pose and expression of $D$, respectively.}
   
\label{fig:vae_sup}
\end{figure}

As shown in Fig. \ref{fig:vae_sup}, we present additional VAE interpolation of expression visualization results, where $D$ provides the pose and expression information of the face to animate $S$.

\begin{table}[t]
 \caption{Quantitative results of ablation study on VoxCeleb \cite{nagrani2017voxceleb}.}
\centering
\resizebox{\textwidth}{!}{%
\begin{tabular}{llllllllllll}
    \toprule
\multirow{2}{*}{Method} & \multicolumn{7}{c}{Same-identity}  & \multicolumn{4}{c}{Cross-identity} \\ \cmidrule(r){2-8} \cmidrule(r){9-12}
  & LPIPS $\downarrow$ & CSIM $\uparrow$  & PSNR $\uparrow$  & APD $\downarrow$ & AKD $\downarrow$ & AED $\downarrow$ & FID $\downarrow$ & CSIM $\uparrow$ & APD $\downarrow$   & AED $\downarrow$ & FID $\downarrow$   \\ \midrule
\textbf{Base model}     
& 0.110     & 0.971   & 22.694  & 0.020  & 1.621 & 0.0250 & 16.781  & 0.925    & \textbf{0.041} & 0.0764 & 88.680  \\ 
\textbf{w/ $\mathcal{L}_{Exp}$ }
& 0.111    & 0.971   & 22.390  & 0.021  & 1.684 & 0.0256 & 14.681  & 0.930    & 0.045 & 0.0760 & 84.108  \\ 
\textbf{w/ $\mathcal{L}_{Exp} \& \mathcal{L}_{C}$ }     
& 0.112    & 0.970   & 22.480  & 0.022  & 1.742 & 0.0256 & 16.232  & \textbf{0.932}    & 0.047 & 0.0760 & 79.960  \\ 
\textbf{w/ $\mathcal{L}_{Exp}, \mathcal{L}_{C} \& \mathcal{L}_{M}$ } 
& \textbf{0.106}     & \textbf{0.974}   & \textbf{22.898}  & \textbf{0.017}  & \textbf{1.476} & \textbf{0.0241} & \textbf{13.265}  & 0.925    & 0.042 & \textbf{0.0759} & \textbf{77.445}  \\ 
    \bottomrule
    \end{tabular}%
  }
\label{table:ablation}
\end{table}

\subsection{Abalation Study}
We are interested in $\mathcal{L}_{Exp}$, $\mathcal{L}_{C}$ and $\mathcal{L}_{M}$ and conduct ablation studies to evaluate its effectiveness. The quantitative results are shown in Tab. \ref{table:ablation}.
We can observe that the metrics in the second row indicate that under the influence of $\mathcal{L}_{Exp}$, the same-identity-driven metrics are basically the same, except for the improvement in FID. However, in cross-identity-driven metrics, AED and CSIM achieve effective improvements, suggesting that $\mathcal{L}_{Exp}$ can enhance the model's identity preservation ability. The metrics in the third row indicate that $\mathcal{L}_{C}$ can further improve the identity preservation ability in cross-identity-driven tasks, which is consistent with the qualitative ablation study results in the manuscript. After applying $\mathcal{L}_{M}$, our model's metrics achieve significant improvements. In the same-identity-driven task, better LPIPS, PSNR, and AKD can be obtained, representing effective improvements in the model's reconstruction ability and expression transfer ability. The increase in APD signifies the enhancement of the model's pose transfer ability, while the improvements in AED and CSIM characterize the maintenance of identity information. In the cross-identity-driven task, the best AED is also achieved. Simultaneously, in both tasks, the combined use of the three losses allows us to achieve the best FID, which represents image quality.

\section{Social Impact} \label{social} 
% There is no direct negative social impact of this work, as it aims to enhance user experience in various applications, including remote consoles, video conferencing, and online customer service. However, we should also prevent this face generation technology from being used in other illegal scenarios.

% Face animation generation technologies have many applications, such as remote control consoles, video conferencing, and gaming. They enhance user experiences and promote immersive interactions. However, misuse can lead to negative social impacts, like deepfake videos and concerns about privacy and consent.

This work primarily focuses on improving face animation generation, which can contribute positively to a wide range of applications such as remote control consoles, video conferencing, online customer service, virtual reality, gaming, and digital content creation. By enhancing the user experience in these areas, it can lead to more immersive and realistic interactions, fostering better communication and collaboration.
However, it is essential to consider the potential negative social impacts that may arise from the misuse of such face animation generation. For instance, the improved face animation could be exploited to create deep fake videos or images, which can be used for spreading misinformation, identity theft, or other malicious purposes. Additionally, concerns about privacy and consent may arise when using individuals' facial data without their permission.

\section{Limitations} \label{limit}
% Since MMFA uses a 3D keypoint estimation network, the model requires larger training resources than 2D keypoint-based models such as FOMM \cite{siarohin2019first}.
% In the future, we plan to explore the possibility of streamlining the network architecture to facilitate rapid training and inference.

Due to the use of 3D keypoint estimation networks and dense motion estimation networks with 3D convolution operations in MMFA, the model requires more training resources compared to 2D keypoint-based models (such as FOMM \cite{siarohin2019first}). Moreover, in same-identity expression driving, 3D keypoint-based methods do not show superiority over the 2D keypoint-based methods, although they have better cross-identity preservation capabilities and image generation quality. In the future, we plan to explore the possibilities of simplifying or optimizing the network architecture to facilitate faster training and inference, and investigate the combination of 2D and 3D keypoints to achieve more realistic animation driving.

%%%%%%%%%%%%%%%%%%%%%%%%%%%%%%%%%%%%%%%%%%%%%%%%%%%%%%%%%%%%

\end{document}